\documentclass{article}

\usepackage{times}
\usepackage{amsmath,amsthm,amssymb}
\usepackage{tabularx}
\usepackage{floatrow}
 
\DeclareMathOperator*{\argmin}{arg\,min} 

\usepackage{graphicx} 
\usepackage{subfigure} 

\usepackage{natbib}

\usepackage{algorithm}
\usepackage{algorithmic}

\usepackage[belowskip=5pt,aboveskip=5pt]{caption}

\usepackage{titlesec}
\titlespacing*{\section}{0pt}{0.8\baselineskip}{\baselineskip}

\usepackage{hyperref}

\usepackage[accepted]{icml2017}

\usepackage{array}

\icmltitlerunning{Conditional Density Estimation with Bayesian Normalising Flows}
\begin{document} 
\setlength{\abovedisplayskip}{3pt}
\setlength{\belowdisplayskip}{3pt}

\graphicspath{{figs/}}
\twocolumn[
\icmltitle{Conditional Density Estimation with Bayesian Normalising Flows }]



\icmlsetsymbol{equal}{*}

\begin{icmlauthorlist}
\icmlauthor{Brian L. Trippe}{cam,mit}
\icmlauthor{Richard E. Turner}{cam}

\end{icmlauthorlist}

\icmlaffiliation{mit}{Massachusetts Institute of Technology, Cambridge, USA}
\icmlaffiliation{cam}{University of Cambridge, Cambridge, United Kingdom}

\icmlcorrespondingauthor{Brian L. Trippe}{btrippe@mit.edu}
\icmlcorrespondingauthor{Richard E. Turner}{ret26@cam.ac.uk}

\icmlkeywords{Normalising flows, Bayesian neural networks, Conditional density estimation}

\vskip 0.3in



\printAffiliationsAndNotice{}  

\begin{abstract} 
Modeling complex conditional distributions is critical in a variety of settings.  Despite a long tradition of research into conditional density estimation, current methods employ either simple parametric forms or are difficult to learn in practice.  This paper employs normalising flows as a flexible likelihood model and presents an efficient method for fitting them to complex densities.  These estimators must trade-off between modeling distributional complexity, functional complexity and heteroscedasticity without overfitting.  We recognize these trade-offs as modeling decisions and develop a Bayesian framework for placing priors over these conditional density estimators using variational Bayesian neural networks.  We evaluate this method on several small benchmark regression datasets, on some of which it obtains state of the art performance. Finally, we apply the method to two spatial density modeling tasks with over 1 million datapoints using the New York City yellow taxi dataset and the Chicago crime dataset.
\end{abstract} 

\vspace{-5pt}
\section{Introduction}
Conditional density estimation (CDE) is a general framing of supervised learning problems, subsuming both classification and regression.   While the objective of most supervised learning algorithms is to accurately predict the expected value of a label $y$ conditional on observing associated features $x$, these methods generally have implicit or explicit probabilistic interpretations (e.g. predicting a categorical distribution in classification problems or the mean of a Gaussian predictive distribution in regression problems).  In these cases, the assumed conditional densities take on a simple parametric form; in contrast, this paper is concerned with a more general setting, in which one is interested in estimating more complex conditional distributions.

Modeling complex and heteroscedastic noise distributions is useful in a variety of settings for which the full predictive distribution rather than its mean is of inherent interest or informs subsequent decisions.   For example, in reinforcement learning, we often want to model action-value functions for risky decisions for which rewards are inherently bimodal, or state transition dynamics which may vary significantly from Gaussian \cite{Depeweg2016}. In financial modeling, properly handling heavy tailed distributions may be crucial.  Additionally, in spatial density modeling, we find extremely non-Gaussian distributions over locations of people and events in space.  In each of these cases, we need to predict strongly non-Gaussian, potentially diverse conditional distributions and would like to learn to make these predictions from a finite dataset.

A number of methods exist for performing CDE in its full generality which, in the limit of very large models and datasets, are able to capture arbitrary conditional distributions.  These date back to adaptive mixtures of local experts \cite{Jacobs1991} and mixture density networks \cite{Bishop2015}, which learn a neural network mapping from observed variables to the parameters of a mixture model over labels.  More recent work, arising primarily from approaches in unsupervised learning, includes conditional variants of variational autoencoders and generative adversarial networks \cite{Sohn2015,Mirza2014}, which can sample from complex conditional distributions but do not provide tractable likelihoods.  Another related line of work has explored autoregressive models for density estimation of complex distributions, primarily focusing on images, and has found these to facilitate learning of high dimensional probability densities \cite{Dinh2014,Murray2014,VandenOord2016,Papamakarios2016}.

However, CDE remains a challenging problem, and a particularly daunting one to solve in its general formulation.  The form of the conditional distributions of interest may be arbitrarily complex and be subject to nontrivial heteroscedastic changes throughout the input space, so a general framework must have the capacity to approximate these potentially complex distributions and include the machinery to infer them from data.  Of particular concern to flexible methods for CDE is the danger of overfitting.  This challenge becomes particularly clear when one considers that we are ultimately interested in inferring conditional distributions from which we have not observed even a single sample.  Even in settings with large datasets, when observed features are high dimensional and diversely distributed, the actual quantities of observed data from relevant conditional distributions may be small.  These cases demand care in order to share statistical strength across samples if we are to model complex distributions without overfitting or underfitting.  As such, methods for CDE face fundamental trade-offs between modeling stationary distributional complexity,  functional complexity, and heteroscedastic changes to the noise distributions and must balance all of these to avoid overfitting or underfitting the available data.

\begin{figure}
\begin{center}
\centerline{\includegraphics[width=1.0\columnwidth]{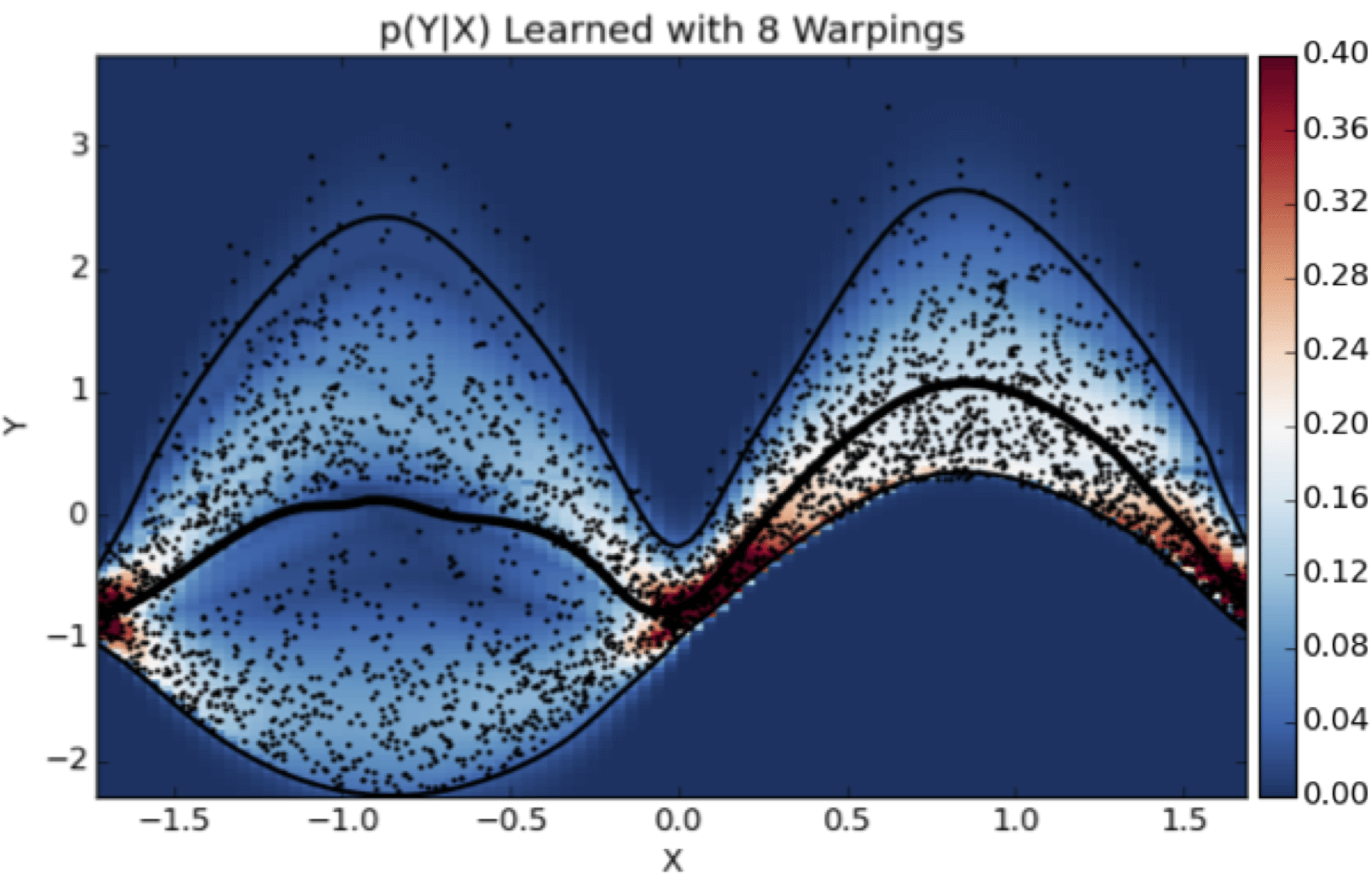}}
\vspace{-20pt}
\caption{Complex, heteroscedastic conditional densities are learned with normalising flows on a toy dataset using the the proposed method (N$=5000$).  Color reflects the predicted conditional probability, $p(y|x)$.  Black lines represent the median and $95\%$ confidence intervals of the conditionals. Best viewed in color.}\label{fig:nflows_toy}
\end{center}
\vspace{-12pt}
\end{figure}

This paper's contribution is twofold.  First, we propose using normalising flows as a flexible likelihood model in conditional density estimation and develop a computationally efficient method for fitting to complex conditional densities (Figure \ref{fig:nflows_toy}).  Second, we confront the trade-offs between modeling distributional complexity and functional complexity by recognizing this as a modeling decision.  To this end we develop a Bayesian framework for CDE with normalising flows using Bayesian neural networks mapping from features, $x$, to the parameters of a normalising flow defining the conditional density, $p(y|x)$.  This allows the explicit placement of priors over conditional distributions defined by normalising flows and over characteristics of their changes throughout the input space such as the extent of heteroscedasticity.

Though exact inference in this class of models is intractable, we use a variational Bayesian approximation to the posterior over the parameters of the neural network.  We validate our method on UCI datasets and achieve state of the art test log likelihoods on several of the test sets.  Next, we tackle multidimensional conditional density estimation problems. Following Larochelle and Murray, we use autoregressive structure to capture complex 2D densities \cite{Larochelle2011}.  Finally we demonstrate the scalability and utility of our method on two spatial density estimation tasks, using the New York City yellow taxi dataset and a Chicago crime dataset, problems where the densities are of inherent sociological interest and can inform more efficient resource allocation.
\begin{figure}
\centering
\includegraphics[width=1.0\columnwidth]{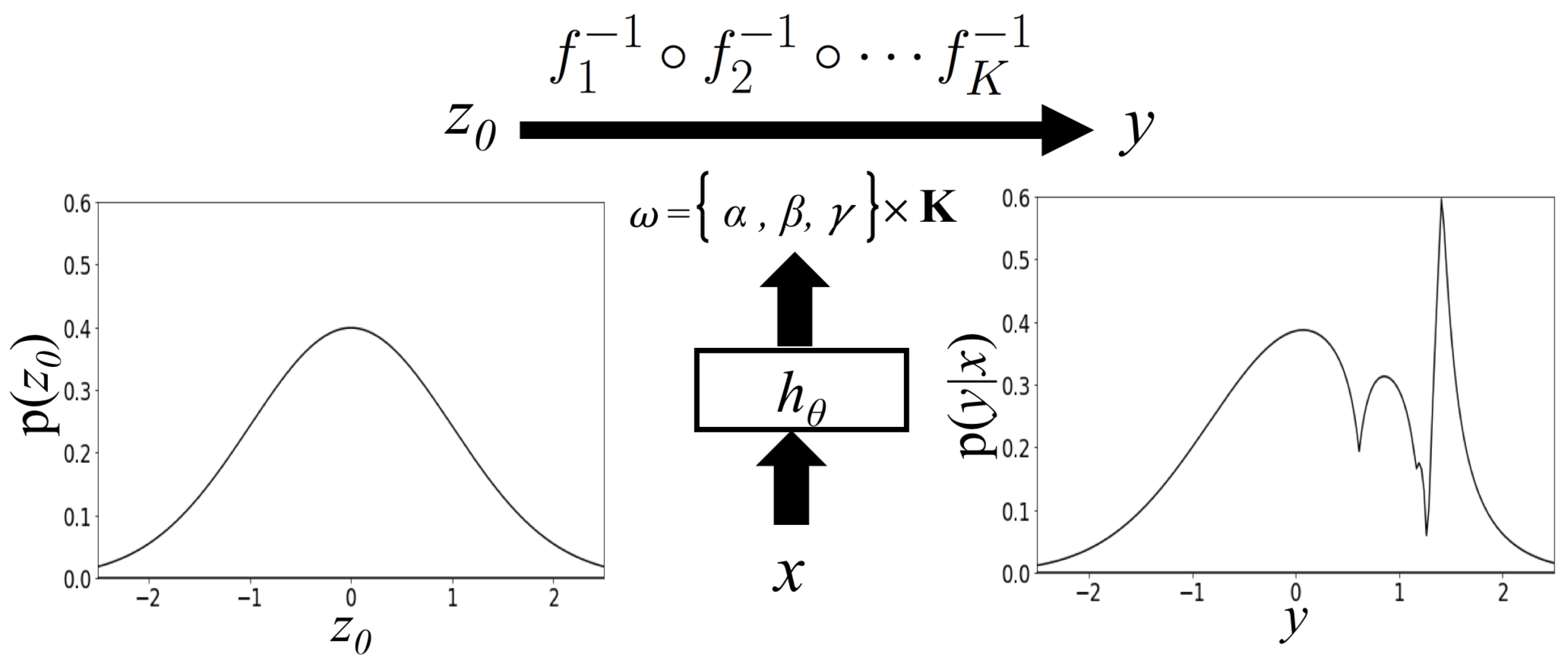} 
\vspace{-10pt}
\caption{A normally distributed random variable passed through a $K-$stage normalising flow gives a complex derived distribution, $p(y=f^{-1}(z)|x)$. In CDE the parameters of the flow are $\omega=h_\theta(x)$.}
\label{fig:nflows_cde}
\vspace{-20pt}
\end{figure}

\begin{table*}
\vspace{-7pt}
\caption{Bayesian normalising flows in relation to some common methods for conditional density methods}
\centering
\label{table:cde_methods}
\fontsize{10}{14.0}\selectfont
{\begin{tabular}{|p{4.24cm}|p{3.cm}|p{2.60cm}|p{5.03cm}|}
\hline
\textbf{Method} & Distribution & Input Dependence & Inference Scheme \\
\hline
Least Squares Regression & Gaussian &$h(x)=\theta^\mathrm{T}x$&$\theta=(X^\mathrm{T}X)^{-1}X^\mathrm{T}Y$\\
\hline
Generalized Linear Models&Exponential Family&$h(x)=\theta^\mathrm{T}x$&Iteratively reweighted least squares\\
\hline
Neural Network Classification&Categorical&$h(x)=\mathrm{NN}_\theta(x)$&Stochastic gradient descent \\
\hline
Mixture Density Networks&Mixture of Gaussians&$h(x)=\mathrm{NN}_\theta(x)$&Stochastic gradient descent \\
\hline
\textbf{This Paper}&Normalising Flows&$h(x)=\mathrm{NN}_\theta(x)$&Stochastic variational inference \\
\hline
\end{tabular}}
\vspace{-15pt}
\end{table*}

\vspace{-5pt}
\section{Conditional density estimation - notation and terminology}
\vspace{-5pt}
Let $\mathcal{D}=\{x_i,y_i\}_{i=1}^N$ be a dataset of observations $(x_i \in X, y_i \in Y)$ sampled i.i.d. from some joint distribution $p(x,y)$.  Conditional density estimation (CDE) refers to the problem of modeling the conditional, $p(y|x,\mathcal{D})$.  In particular, parametric methods for CDE propose a class of densities, $\{p, \omega \in \Omega\}$, and a class of functions, $h$, indexed by $\theta \in \Theta $, and use $\mathcal{D}$ to choose, $h_\theta: X \rightarrow \Omega, x_i \mapsto \omega_i$, which is then used to model $p(y_i|x_i)$ as $p(y_i|\omega_i=h_\theta(x_i))$.  The choice of $\Omega$ determines the sort of distributional complexity which may be learned from $\mathcal{D}$, and $\Theta$ sets the array of input dependent changes which can be learned, ranging from simple global translations of a stationary predictive distribution to complex heteroscedastic behavior.

Methods for CDE are defined entirely by choices of $\Omega$, $\Theta$ and an inference procedure, which prescribes an objective and a learning algorithm that dictate how to choose $\theta$ (Table \ref{table:cde_methods}).  For example, linear regression models define a linear mapping from $x$ and $\theta$ to $\omega$, where $\omega$ defines an exponential family likelihood.  Ordinary least squares additionally specifies that $p(y|\omega)$ is Gaussian, and defines a maximum likelihood objective with an analytic form to find $\theta$.  Neural network classifiers use $\theta$ to define a neural network and $\omega$ to be the softmax-transformed output defining a categorical distribution over labels, such that $p(y|\omega=h_\theta(x))=\mathrm{Cat}(y|\omega)$, and typically use stochastic gradient descent to find $\theta$.

In this work, we investigate using normalising flows \cite{Tabak2013} as a likelihood model in CDE. In particular we use an adaptation of radial flows which consist of warpings with $3$ free parameters each, and use a neural network to predict these parameters as a function of each $x_i$ \cite{Rezende2015}.  To avoid overfitting with these flexible choices of $\Omega$ and $\Theta$, we perform variational inference (VI) over, $\theta$, the parameters of the neural network (Figure \ref{fig:nflows_cde}).

\vspace{-5pt}
\section{Normalising flows}\label{sec:cde_nflows}
\vspace{-5pt}
A normalising flow is a mapping between probability densities defined by a differentiable, monotonic bijection between the spaces in which they live \cite{Tabak2013}.  These mappings are composable, and a series of relatively simple invertible transformations can be used define more complex transformations.  Rezende and Mohammed introduced two families of parametric transformations, and showed that a series of these transformations could warp a standard Gaussian base distribution into rich approximate posteriors \cite{Rezende2015}.  In particular, by mapping a random variable $z_0$ through a $K$-stage normalising flow, $f=(f_1,f_2, \dots, f_K)$, we define a transformed variable, $z_k$, and its derived distribution:
\begin{multline}\label{eqn:rezende_flow_definition}
z_K = f_K(f_{K-1}( \dots f_1 (z_0))) \ \ \ \ \ \mathrm{ and}  \\
\mathrm{ln}\ p(z_K) = \mathrm{ln}\ p(z_0) - \sum_{k=1}^K \mathrm{ln} \frac{d f_k(z_{k-1})}{d z_{k-1}} 
\end{multline}
Where $p(z_0)$ is defined to be a standard normal base distribution and each $f_k$ is a simple, monotonically increasing function which has a closed form with an easy to calculate derivative.  In particular, we will soon consider `radial flows' \cite{Rezende2015}.

\vspace{-7pt}
\subsection{Inverted normalising flows}
As this formulation of normalising flows was developed for Monte Carlo variational inference, it is optimized for efficiently drawing samples and calculating likelihoods and is not immediately suitable to CDE.  As recognized by \citet{Papamakarios}, the utility of this class of transformations to conditional density estimation is limited by the inefficiency of evaluating the likelihood of externally provided data, which requires inverting each $f_k$.  We overcome this limitation by reversing the direction of the normalising flows, defining the forward mapping from the target, $y$, to the base distribution:
\begin{multline}\label{eqn:de_flow}
z_0 = f_1 (f_2 ( \dots f_K (y))) \ \ \ \ \ \mathrm{ and} \\ 
\mathrm{ln}\ p(y) = \mathrm{ln}\ p(z_0) + \sum_{k=1}^{K-1} \mathrm{ln} \frac{d f_k(z_k)}{d z_k}+ \mathrm{ln} \frac{d f_K(y)}{d y}
\end{multline}

Where $f_K: y \mapsto z_{k-1}$, and for $k<K$, $f_k: z_k \mapsto z_{k-1}$ and $p(z_0)$ is defined to be a standard normal base distribution.  Notably, the sign of the log gradient terms is flipped in equation \ref{eqn:de_flow} relative to equation \ref{eqn:rezende_flow_definition}, as the derivative of the inverse transformation is the inverse of the derivative of the forward mapping.  This inversion of the direction of parameterization is necessary for the application of this class of normalising flows to CDE as it allows the density of data to be evaluated trivially in constant time.  Without this inversion, pointwise evaluation of the density requires an inefficient inversion with complexity which is logarithmic in the desired precision.  In the conditional setting, the parameters defining each $f_k$ are outputs of $h_\theta(x_i)$, thereby yielding a different conditional distribution for each $x_i$.  

\vspace{-7pt}
\subsection{A new parameterization of radial flows}
\vspace{-5pt}
An additional challenge to applying NFs to CDE that is not critical when using them for inference is the ease of overfitting.  While the variational objective prevents overfitting of learned variational approximations, naive approaches to CDE such as maximum likelihood estimation are prone to overfitting given flexible models \cite{Bishop2006}.

The Bayesian framework provides a compelling approach for avoiding overfitting through modeling parameter uncertainty.  However, the effective use of Bayesian methods requires reasonable priors, and it is not immediately clear how to reason about priors over normalising flows.  In this vein, we developed an alternative parameterization of radial flows \cite{Rezende2015} with which we can more readily express priors over distributions:
\begin{equation}\label{eqn:new_radial_flow}
f(z) = z + \frac{\alpha\beta(z-\gamma)}{\alpha + |z-\gamma|}
\end{equation}
Where the parameters are $\{  \alpha, \beta, \gamma \in{\rm I\!R}\} $. 

\begin{figure}[!ht]
\begin{center}
\centerline{\includegraphics[width=1.0\columnwidth]{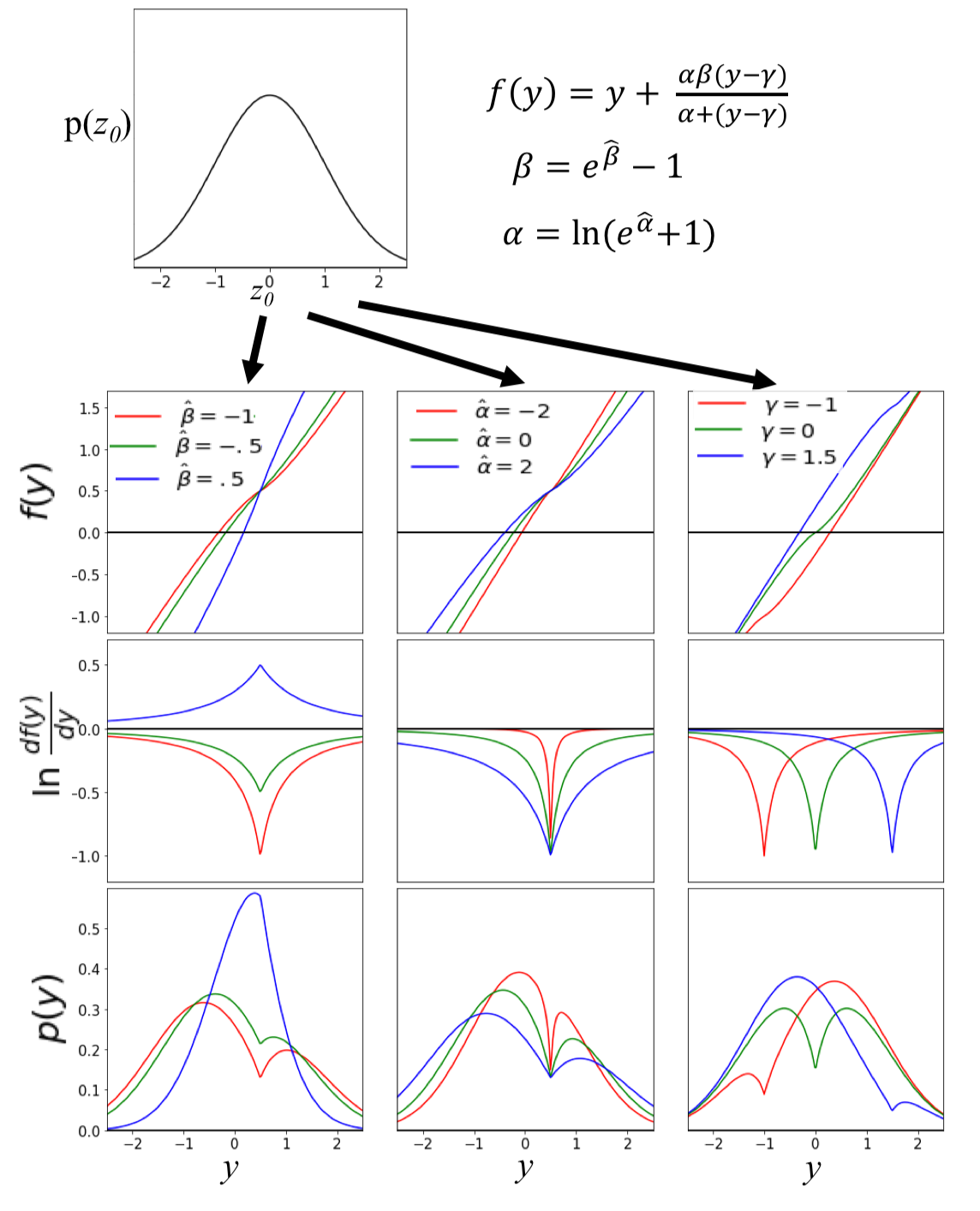}}
\vspace{-10pt}
\caption{The radial flow is a $3-$parameter transformation for defining complex densities by warping a normal base density. Left: $\hat \beta$ defines the maximum of $\mathrm{ln} \frac{df(y)}{dy}$. Middle: $\hat \alpha$ sets the range of the distortion. Right: $\gamma$ sets the location of distortion.}\label{fig:nflows_de}
\end{center}
\vspace{-17pt}
\end{figure}

\begin{figure*}[!ht]
\floatbox[{\capbeside\thisfloatsetup{capbesideposition={right,top},capbesidewidth=4cm}}]{figure}[\FBwidth]
{\caption{A manifold representing probability densities sampled from different priors over a 10-stage normalizing flow.  The same random seed is used to interpolate between different choices of priors to demonstrate the impact of different choices of priors on the resultant distribution.  The variance in maximum magnitudes of the distortions are controlled by $\sigma_{\hat \beta}$, which varies across the densities within each subplot.  The sharpness of the distortions is controlled by $\mu_{\hat \alpha}$, which varies from sharpest to smoothest across the columns.  The variance of the sharpness of the distortions is controlled by $\sigma_{\hat \alpha}$, and is increased in successive rows.    The remaining parameters are fixed at $\sigma_{\hat \beta}=1.0$, $\mu_z=0.0$ and $\mu_{\hat \beta}=0.0$.  Best viewed in color.}\label{fig:sampled_distributions}}
{\includegraphics[width=0.72\textwidth]{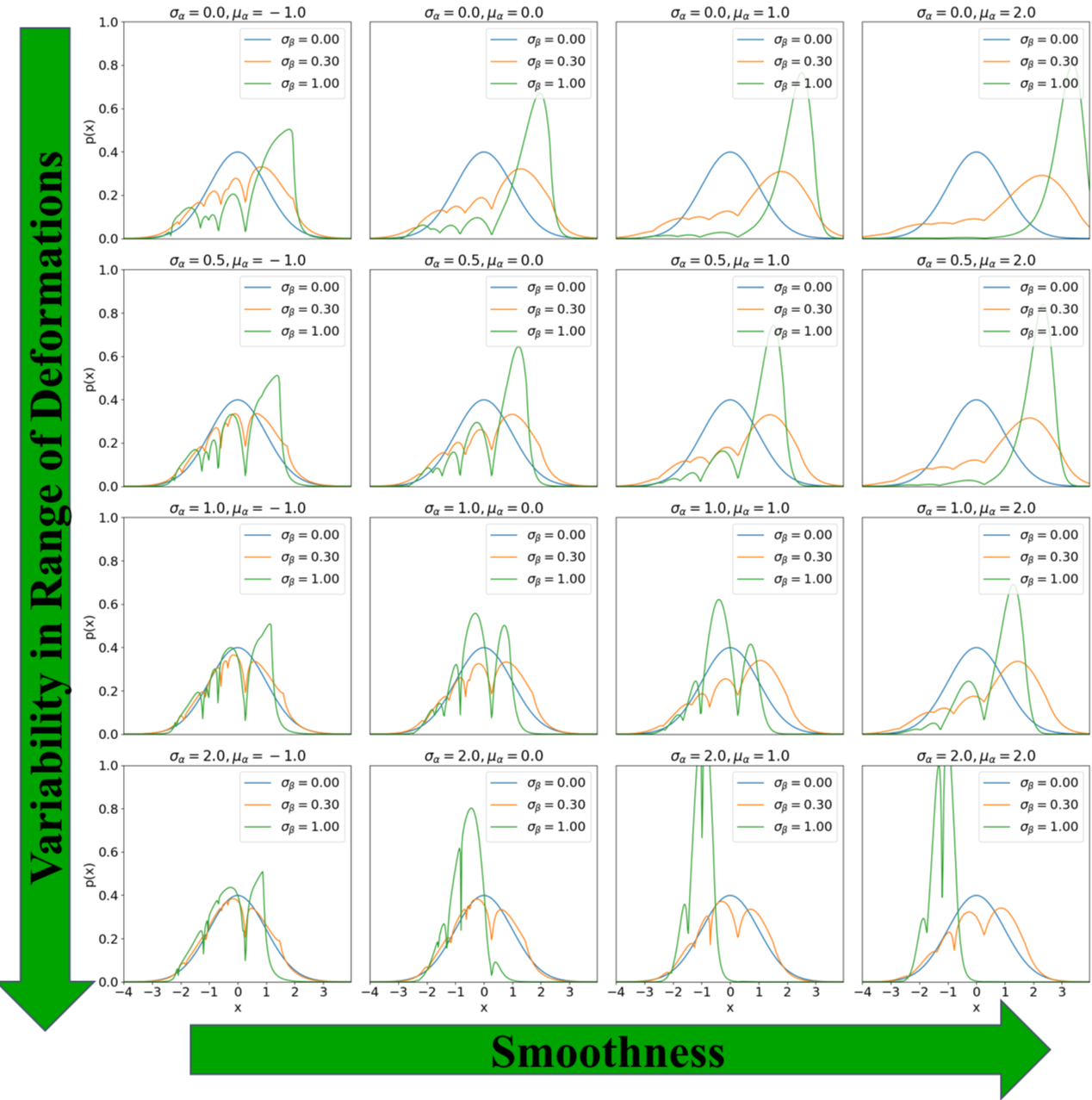}}
\centering
\vspace{-12pt}
\end{figure*}

We can gain intuition into how this function shapes a probability density by examining its gradient (derivation in supplementary equation \ref{eqn:rad_flow_gradient}):
\begin{equation}\label{eqn:new_flow_gradient}
\frac{d f(z)}{dz} = 1+\frac{\alpha^2\beta}{(\alpha + |z - \gamma|)^2}
\end{equation}
Looking closely at $f$ and its derivative, we see that the warping varies from the identity to the greatest extent when $z=\gamma$, where  $f^\prime (\gamma) = 1+\beta$.  The first shape parameter, $\alpha$, controls how quickly $\frac{d f}{d z}$ decays to $1$ away from $\gamma$; large $\alpha$s define broad distortions whereas small, positive $\alpha$s define sharper distortions (Figure \ref{fig:nflows_de}, supplementary figure \ref{fig:sampled_distributions}).  As $\alpha \rightarrow 0$, $f$ collapses to the identity function, and $\alpha<0$ breaks the monotonicity requisite for equation \ref{eqn:de_flow} to reflect the derived distribution, $\mathrm{ln} p(y)$.  We enforce this monotonicity by parameterizing $\alpha$ as a softplus transformed unconstrained parameter, as $\alpha = \mathrm{ln}(\mathrm{exp}(\hat  \alpha) + 1)$, where $\hat \alpha \in \mathbb{R}$.

The second shape parameter, $\beta$, controls the magnitude and direction of the maximum distortion.  If $\beta>0$ then $\mathrm{ln} \frac{df(z)}{dz}>0$, corresponding to a compression of the base density, whereas if $\beta<0$ then $\mathrm{ln} \frac{df(z)}{dz}<0$ which thins the base density around $\gamma$ (Equation \ref{eqn:de_flow}).  
When $\beta<-1$, $f$ is again no longer monotonic.  Accordingly, we ensure monotonicity by enforcing $\beta \geq -1$ with the parameterization $\beta = \mathrm{exp}(\hat \beta) -1$, where $\hat \beta \in \mathbb{R}$.  Of a number of possible parameterizations, this one is particularly appealing for two reasons.  First, it is independent of first shape parameter $\alpha$\footnote{This is not the case for previous paramterisations \cite{Rezende2015}.} and second, the untransformed parameter, $\hat \beta$, is equal to the maximum magnitude of the log derivative of the function it parameterizes.  \emph{As a result, priors we place on $\hat \beta$ are on the maximum change in log density of points in the base distribution:}
\begin{equation}
\mathrm{ln} \frac{d f}{d z}(\gamma) = \mathrm{ln}\big( 1+ \beta \big) =  \hat \beta 
\end{equation}
The structure of this parameterisation of radial flows makes it possible to reason about the relationship between the priors we place on the parameters of the radial flows and the probability densities they define (Figure \ref{fig:nflows_de} and supplementary figure \ref{fig:sampled_distributions}).

\vspace{-5pt}
\section{Bayesian conditional density estimation with normalising flows}
\vspace{-5pt}
In this section we introduce a Bayesian approach to performing CDE using normalising flows as a likelihood model.  We show how to place priors over conditional density estimators, then demonstrate how to accomplish this using Bayesian neural networks and close with a method for approximate inference.

\begin{figure*}[!ht]
\floatbox[{\capbeside\thisfloatsetup{capbesideposition={right,top},capbesidewidth=4cm}}]{figure}[\FBwidth]
{\caption{The characteristics of conditional density estimators sampled from different priors vary with the prior parameters.  Each panel is a heatmap depicting the conditional density estimator defined by an MLP mapping to the parameters of a $5$-stage normalising flow.  The same random seed for each sample, and the reparameterization trick is used to interpolate between different Gaussian priors.  Moving left to right, we increase the prior standard deviation over the parameter $\hat \beta$, which controls the magnitude of the warping.  Moving top to bottom, we increase the value of the parameter $\lambda$, which controls the extent of heteroscedasticity (with larger values reflecting greater heteroscedasticity).  Best viewed in color.}\label{fig:cd_manifold_mid_ls}}
{\includegraphics[width=0.68\textwidth]{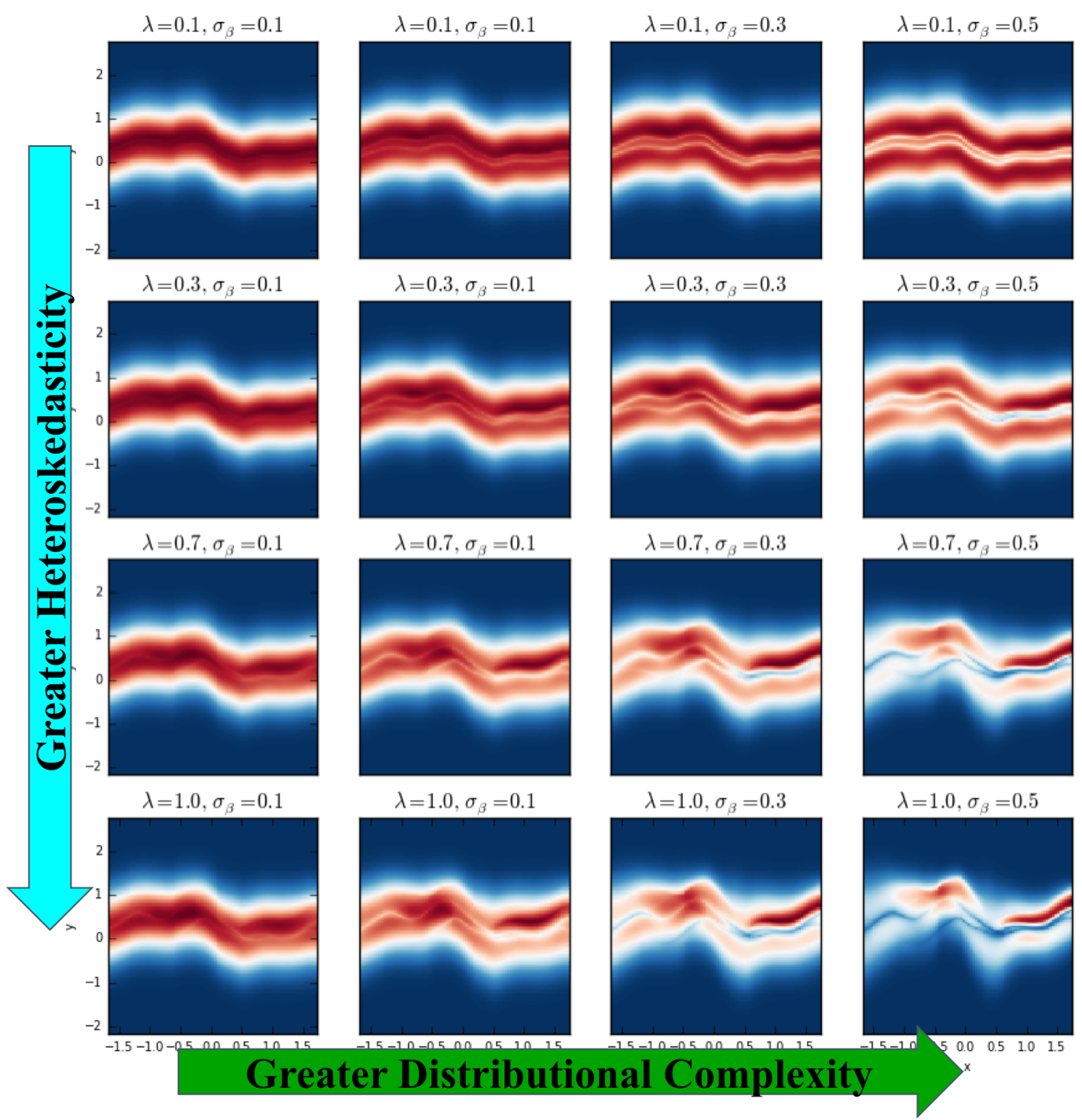}}
\centering
\vspace{-17pt}
\end{figure*}

\subsection{Placing priors over conditional density estimators}
\vspace{-5pt}
In function approximation, inferences about the values functions take on at unseen points arise from assumptions about the function's continuity and smoothness.  In particular, we often assume nearby inputs have `similar' outputs.  A notion of similarity of inputs is crucial to making reasonable inferences about a function.  When we choose a model class, we are defining what it means for an $x_i$ and $x_j$ to be close to one another, be it through the choice of a covariance function in a Gaussian process or the architecture of a deep neural network.  In CDE, we generalize this assumption of similarity to the conditional distribution.  We now assume that \textit{for `similar' observed variables, the corresponding conditional distributions will be `similar' to one another}:
$$
x_i \approx x_j \implies p(y|x_i) \approx p(y|x_j)
$$
This assumption in CDE, clearly begs the additional question; What does it mean for distributions to be `similar'?  We should ideally choose a definition of similarity which reflects our understanding of the problem we are solving. 

For example, in estimating the demand for taxis in a city conditioned on the time of day, we expect to see significant shifts in which areas of the city have greatest density, but we might expect that centers of  density will generally exist in the same places (e.g. around major train stations and highly populated areas) and never in others (e.g. rivers).  In contrast, if we are considering the probability of a worker's wage conditioned on the year, we might expect some characteristics of this distribution to be conserved across years (e.g. heavy upper tails or sharp peaks at minimum wages) but we also expect global translations of this distribution as a result of slowly varying trends, such as inflation, changing minimum wage or economic growth.  Our priors about how $p(y|x)$ changes as $x$ changes should inform how we define this notion of distance between distributions and infer a conditional density estimator.

Questions about how to define distributional similarity are difficult and important ones to answer.  However, we leave this to future work and fall back on a straightforward, if dissatisfying, notion of similarity between distributional defined with normalising flows - closeness in parameter space.  By placing a prior on $\theta$ that reflects a belief that $h_\theta$ is smooth and slowly varying, we encode a belief that the conditional distributions change slowly throughout the space as well.

\subsection{Bayesian neural networks and normalising flows}
\vspace{-5pt}
The usage of similarity in parameter space as the underlying notion of distributional similarity sheds additional light on the importance of choice of parameterization towards our ability to place good priors on conditional distributions.  For the parameterization of radial flows in equation \ref{eqn:new_radial_flow}, distance in parameter space has an intuitive interpretation when considering distance in the space of distributions.  By placing priors how the $\hat \beta$s change throughout the input space, we directly place priors on how the maximum expansion and contraction of the base distribution induced by each stage of the normalising flow will change.  Priors how $\hat \alpha$ and $\gamma$ vary in turn act as priors on how the sharpness and center points of the distortions will vary.  Tuning our of priors on the variances of these parameters relative to one another translates directly into priors over the shifts in conditional distributions we believe best explain data we see.

Bayesian neural networks provide a powerful tool for placing tunable priors over functions by placing priors over the parameters of a neural network \cite{MacKay1992,Neal1995}.  In particular, by placing Gaussian priors with varying means and variances over the weights and biases connecting into the output units defining each normalising flow parameter, on can encode prior beliefs about the corresponding characteristics of CDEs.  Naturally, the variance of a zero-mean prior over the parameters which determine the value of $\hat \beta$, $\sigma_{\hat \beta}$, defines the extent of non-Gaussianity;  Large values of $\sigma_{\hat \beta}$ lead to complex predictive distributions and as $\sigma_{\hat \beta} \rightarrow 0$ we recover the Gaussian base distribution.

Perhaps the clearest motivation for defining a notion of distributional similarity through specification of priors is the trade-off between modeling complexity of conditional distributions and the complexity of their changes throughout the input space.  Models for CDE are often split into homoscedastic and heteroscedastic models, where homoscedastic models have a stationary noise distribution and heteroscedastic models have varying noise distributions throughout the input space.  However, by choosing $h_\theta$ to be a Bayesian neural network, we can smoothly interpolate between homoscedastic and heteroscedastic models by adapting our prior on the weights mapping from the hidden layer to the parameters of the normalising flows, we do this by introducing an additional hyperparameter, $\lambda$, which defines a multiplicative scaling of the final layer weights (but not biases).  For example, when these weights are small, the parameters of conditional distributions will vary only slightly from their biases.  The resulting functions have the same length scale for changes in noise structure, but the magnitude of the deviations is tuned by the hidden-to-output weights.  One can get a better sense of the beliefs different priors express about conditional density estimators by considering samples drawn from these priors (Figure \ref{fig:cd_manifold_mid_ls}).  In this way, we can see how larger values of $\lambda$ give rise to greater heteroscedasticity.

\begin{figure*}[!ht]
\begin{center}
\centerline{\includegraphics[width=0.57\textwidth]{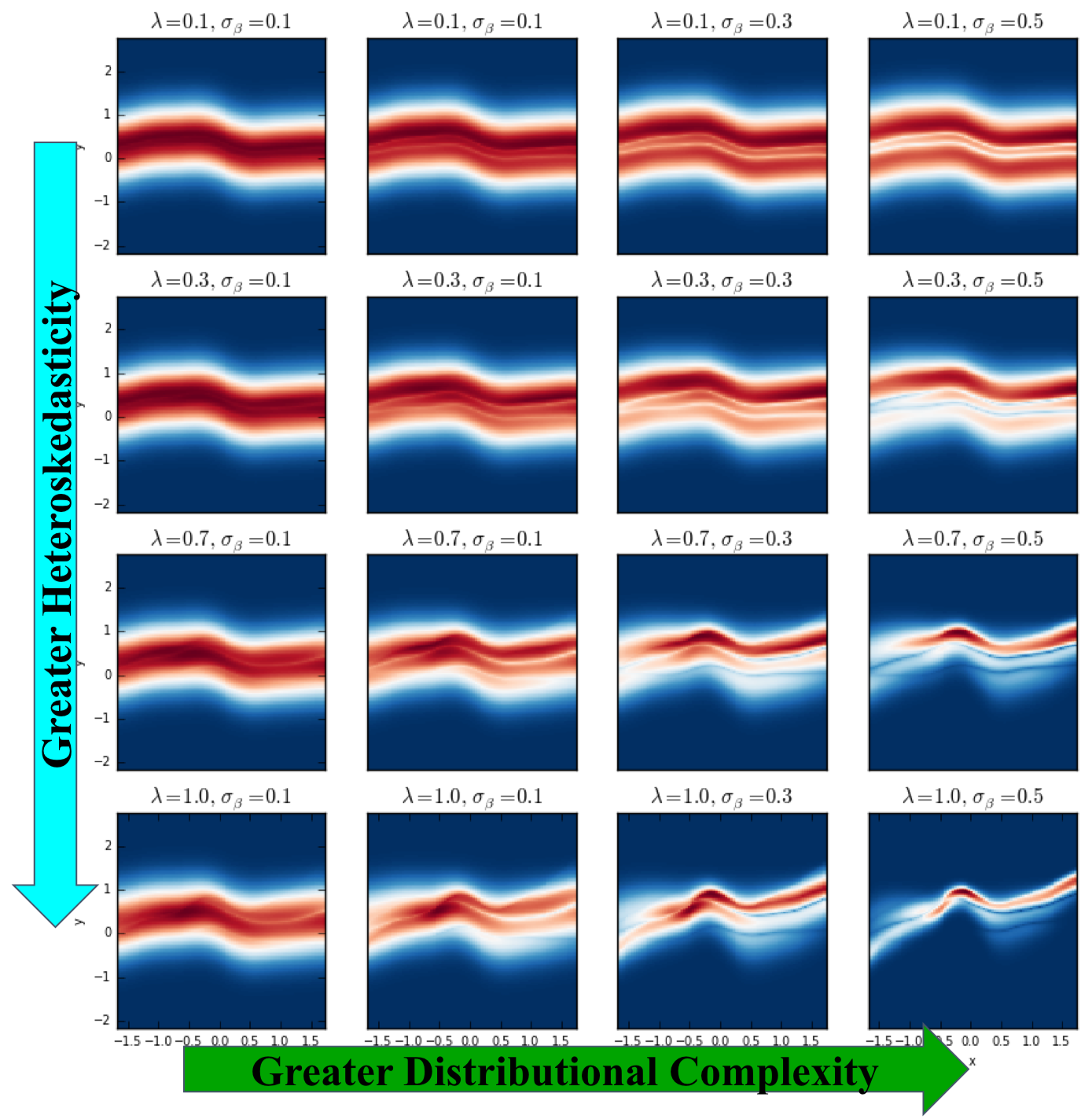}}
\caption{An additional slice of the manifold in Figure \ref{fig:cd_manifold_mid_ls} with a longer length scale.}\label{fig:cd_manifold_slow_ls}
\end{center}

\end{figure*}

Similarly, Bayesian neural networks enable one to set a prior length scale for the changes in the noise distribution (through the prior over input-to-hidden weights \cite{Neal1995}) (Supplementary figure \ref{fig:cd_manifold_slow_ls}).  In the models explored in this paper, we use a single neural network with multiple outputs to define every parameter of the NFs, but note that this expresses strong beliefs about the length scales of the changes in characteristics of these distributions being similar, which may not be a good assumption in general.

\subsection{Inference and hyperparameter selection}\label{sec:inference}
\vspace{-5pt}
Exact posterior inference over the parameters of a Bayesian neural network is intractable.  As such, we turn to variational inference for a tractable approximation.  Variational inference (VI) \cite{Jordan1998} minimises the KL-divergence between an approximation, $q(\theta)$, of the intractable posterior, $p(\theta|\mathcal{D},\alpha)$:
\begin{multline}\label{eqn:vfe}
\argmin_{q \in \mathcal{Q}} \mathrm{KL}[q||p] = \argmin_{q \in \mathcal{Q}} \mathcal{F}(q) \mathrm{,  where}\\
\mathcal{F}(q)=  -\mathbb{E}_{ q(\theta)} \big[\mathrm{log} \  p(Y|X, \theta)\big]+ \mathrm{D_{KL}}\big(q(\theta)||p(\theta|\alpha)\big)
\end{multline}

Where $\mathcal{Q}$ is the variational family, and $\mathcal{F}$ is known as the variational free energy.  
We use a mean field Gaussian approximation which we fit by stochastic variational inference \cite{Graves2011,Blundell2015}.  Our variational approximation uses tied, fixed posterior variances over weights and biases which we generally find to perform more favorably than the untied version \cite{Wang2013,Trippe2017a} and the local reparameterization trick to reduce the variance of the Monte Carlo gradient estimates \cite{Kingma2015}.  Priors over weights and biases are kept as hyperparameters of the model, to be chosen in a problem dependent manner based upon either prior knowledge or a model selection scheme such as Bayesian optimization or cross-validation.  In this way, we have turned prior beliefs about the extent of non-Gaussianity and heteroscedasticity into hyperparameters of our model.

\subsection{Related Work}\label{sec:related_work}
\vspace{-5pt}
Two Bayesian methods for CDE with flexible likelihood models have recently been proposed.   \citet{Papamakarios2016} perform stochastic variational inference in a mixture density network for an application to approximately Bayesian computation and \citet{Depeweg2016} and coauthors use stochastic inputs to a Bayesian neural network for an application to reinforcement learning.  However, while both methods use neural network based models and perform approximate inference over their parameters, neither work discusses considerations of the choice of prior or the consequences thereof to the conditional densities which are learned.  We compare to these methods in the next section.

\vspace{-5pt}
\section{Results on Benchmark Datasets}\label{sec:benchmarks}
\vspace{-5pt}
\begin{figure}
\begin{center}
\centerline{\includegraphics[width=\columnwidth]{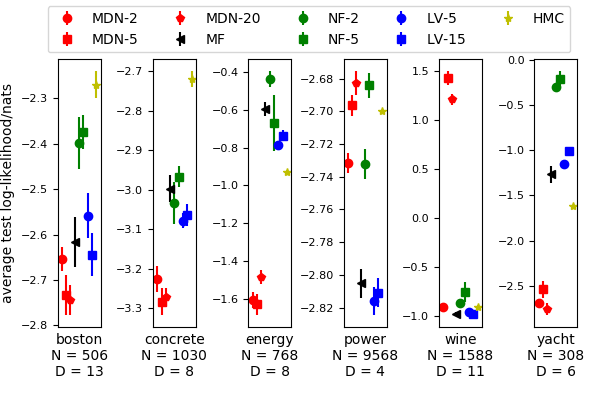}}
\vspace{-20pt}
\caption{Comparison of performance of flexible Bayesian methods for conditional density estimation on six small UCI datasets across 20 train/test splits (Mean$\pm 1$SEM).  Higher is better.}
\vspace{-12pt}
\label{fig:cde_uci_compare}
\end{center}
\vspace{-5pt}
\end{figure}
We evaluated Bayesian normalising flows (NF) on six benchmark UCI regression datasets, comparing against two alternative Bayesian methods for CDE which can approximate arbitrarily complex conditional distributions in the limit of large models; mixture density networks (MDN) and neural networks with latent variable inputs (LV) using Bayesian implementations following \citet{Papamakarios2016} and \citet{Depeweg2016}, respectively.  For each of these three models, we tested two levels of complexity of the predictive distributions (i.e. number of warpings, mixing components and noise samples, respectively), which we chose to be roughly equivalent in expressivity.  We additionally compare to Bayesian neural networks models with homoscedastic Gaussian likelihoods using two approximate inference methods; a mean-field variational approximation (MF) and a sampler using Hamiltonian Monte Carlo (HMC)\footnote{We include the results for HMC as reported by \cite{Bui2016} on these datasets using the same train/test splits.}.  Hyperparameters of all methods were optimized on held out validation sets using Bayesian optimization.  Implementation details of all methods are provided in supplementary section \ref{sec:comparisons}.

Of the methods tested, we see the best overall performance by the more expressive normalising model, NF-5 (Figure \ref{fig:cde_uci_compare}).  The normalising flow based models outperform MF on every dataset except for `energy', on which performance is not significantly different.  Additionally, they yield state of the art performance on two of these datasets, `energy' and `yacht' \cite{Hernandez-Lobato2015,Bui2016,Li2016,Louizos2016}.  We note that this state of the art performance is with a neural network model consisting of a single hidden layer of $50$ hidden units and expect that wider and deeper models can provide even further improvements.

\begin{figure}
\begin{center}
\centerline{\includegraphics[width=1.\columnwidth]{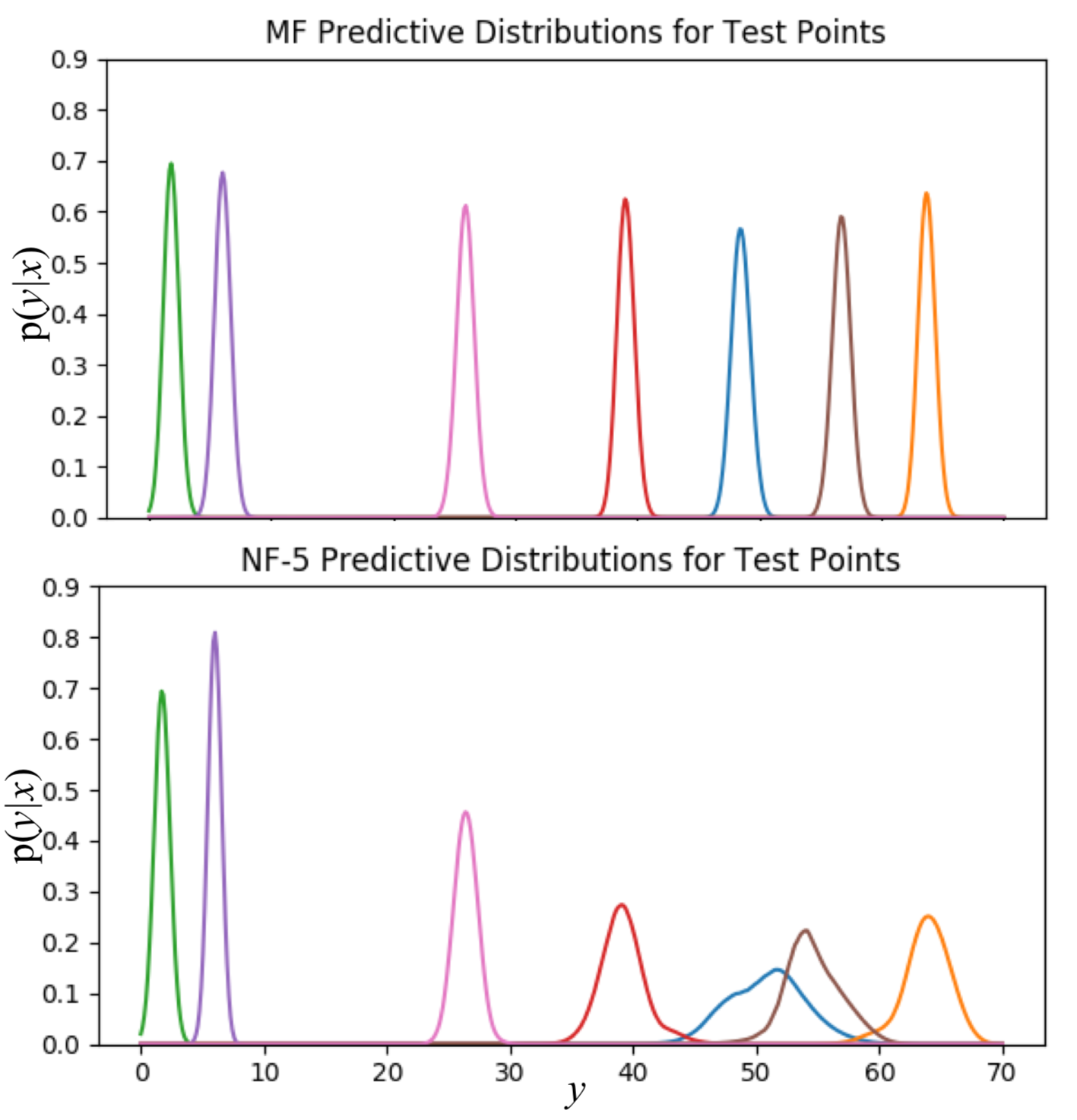}}
\caption{Predictive distributions for MF model and NF-5 model on dataset 'yacht' for 20 test inputs.}
\label{fig:yacht_predictive_dists}
\end{center}
\end{figure}

Normalising flows see the most significant performance increase relative to MF and HMC on `yacht'.  To better understand the source of this performance gain, we looked at the predictions for test points in the first train/test split and found that NF-5 exhibited non-Gaussian predictive distributions with varied noise structure (Supplementary figure \ref{fig:yacht_predictive_dists}).  We believe that capturing this complexity and heteroscedasticity is what allows NF-2 and NF-5 to outperform MF and HMC.

The MDN and LV models perform worse than the NF models on on most datasets.  The exception is the wine quality prediction task, `wine', on which MDN-5 had far superior performance (Figure \ref{fig:cde_uci_compare}).  Upon closer inspection, we found that the labels for this benchmark regression task are ordinal ratings on a $1$ to $10$ scale, and the MDN was able to fit a Gaussian with very small variance to one of these ratings.  As a result this dataset is atypical.

The poor performance of MDN-2 and MDN-5 on the rest of the datasets led us to speculate that these models might be limited by the expressivity of their conditional distributions relative to the normalising flow models.  To test this hypothesis we tested an additional mixture density network with 20 mixing components, which we refer to as MDN-20.  Surprisingly, this model provided better performance than both MDN-2 and MDN-5 on some datasets but performed the same or worse on others.  We suspect that this is due to tied priors variances for all weights and biases failing to optimally capture the trade-offs between modeling distributional complexity and functional complexity.  In particular, we suspect that MDN-5 and MDN-20 are overfitting to stationary distributional complexity on the smallest datasets: `boston' and `yacht'.  Further efforts investigating intelligent placement of priors over mixture density networks may improve the effectiveness of this class of models.

HMC remains the gold standard for inference in BNNs when computational resources are not limiting \cite{Neal1995,Bui2016}.  It significantly outperforms MF on all datasets except for on `energy' and `yacht', on which we suspect the sampler has not completely mixed.  We suspect that accurate Bayesian inference is a major limiting factor in the performance of NF-2 and NF-5 and believe that better inference schemes for conditional density estimators using normalising flows will have similar benefits to those provided to the mean-field variational approximation.

\vspace{-5pt}
\section{Conditional spatial density estimation}\label{sec:nade}	
\vspace{-5pt}
\begin{figure*}[!ht]
\vspace{-7pt}
\floatbox[{\capbeside\thisfloatsetup{capbesideposition={right,center},capbesidewidth=3.5cm}}]{figure}[\FBwidth]
{\caption{Heatmaps depicting learned conditional densities of yellow taxi pick-ups across Manhattan for different fares and tip amounts.  Heat density shows the probability of the pickup location for a given fare and tip amount, $\frac{p(\mathrm{pick-up}|\mathrm{fare},\mathrm{tip})}{d \mathrm{lat}\ d\mathrm{long}}$, and is capped at $1000$.  Best viewed in color.}\label{fig:taxi_fare_tip}}
{\includegraphics[width=0.76\textwidth]{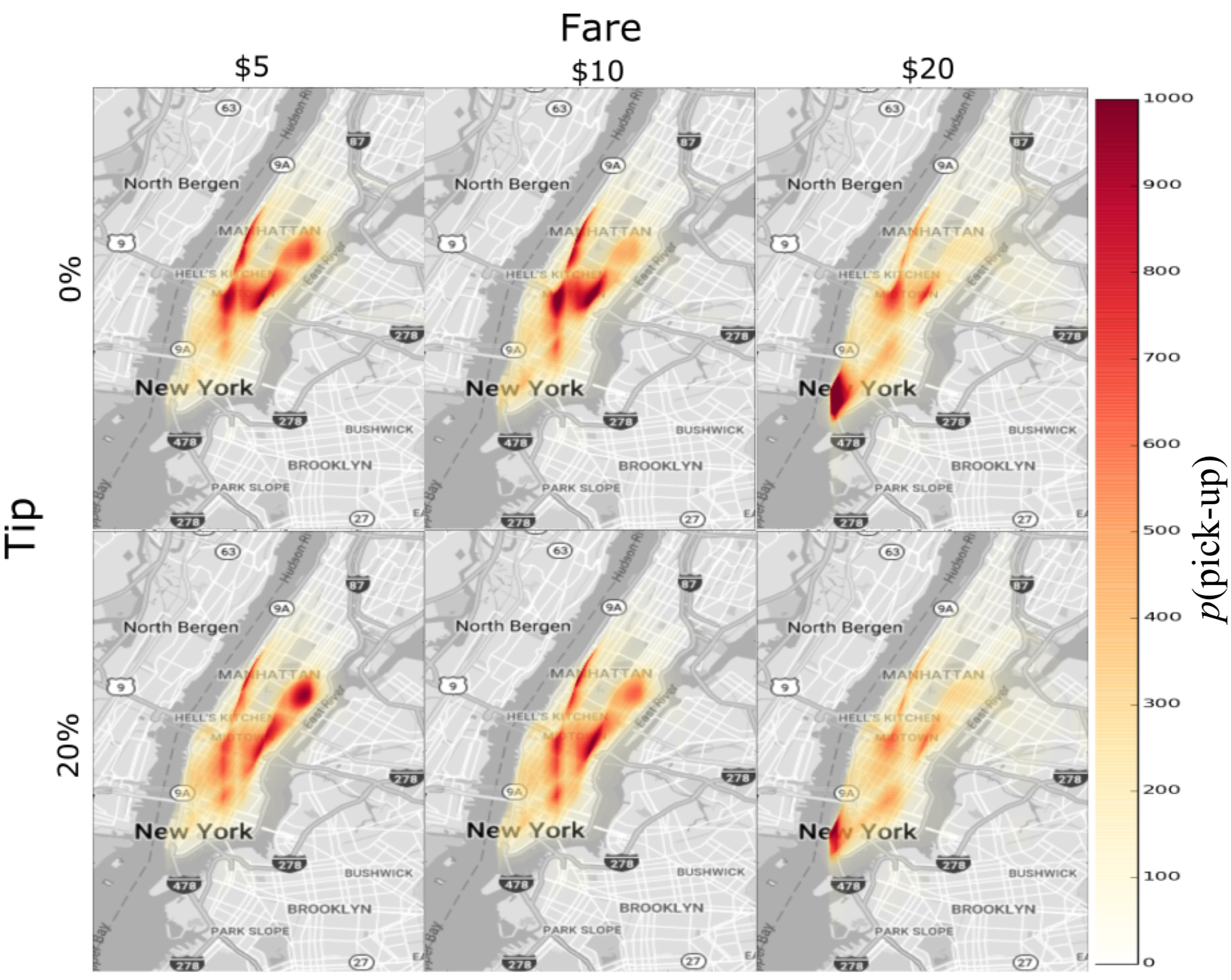}}
\centering
\vspace{-17pt}
\end{figure*}
Conditional spatial density estimation is the problem of predicting 2D distributions over where events occur in space.  In many applications, analytic spatial densities may be of inherent sociological interest or be useful for efficient resource allocation.  In this section we describe an extension of the approach developed in previous sections to capture multidimensional target distributions and then demonstrate its utility on two real-world spatial density estimation tasks, modeling the densities of taxi pickups in New York city in section \ref{sec:spatial} and crime in Chicago in supplementary section \ref{sec:chicago_crime}.  The learned conditional densities provide sociological insights and could inform more efficient resource allocation.

\subsection{Capturing multidimensional predictive distributions}
\begin{figure}
\begin{center}
\centerline{\includegraphics[width=\columnwidth]{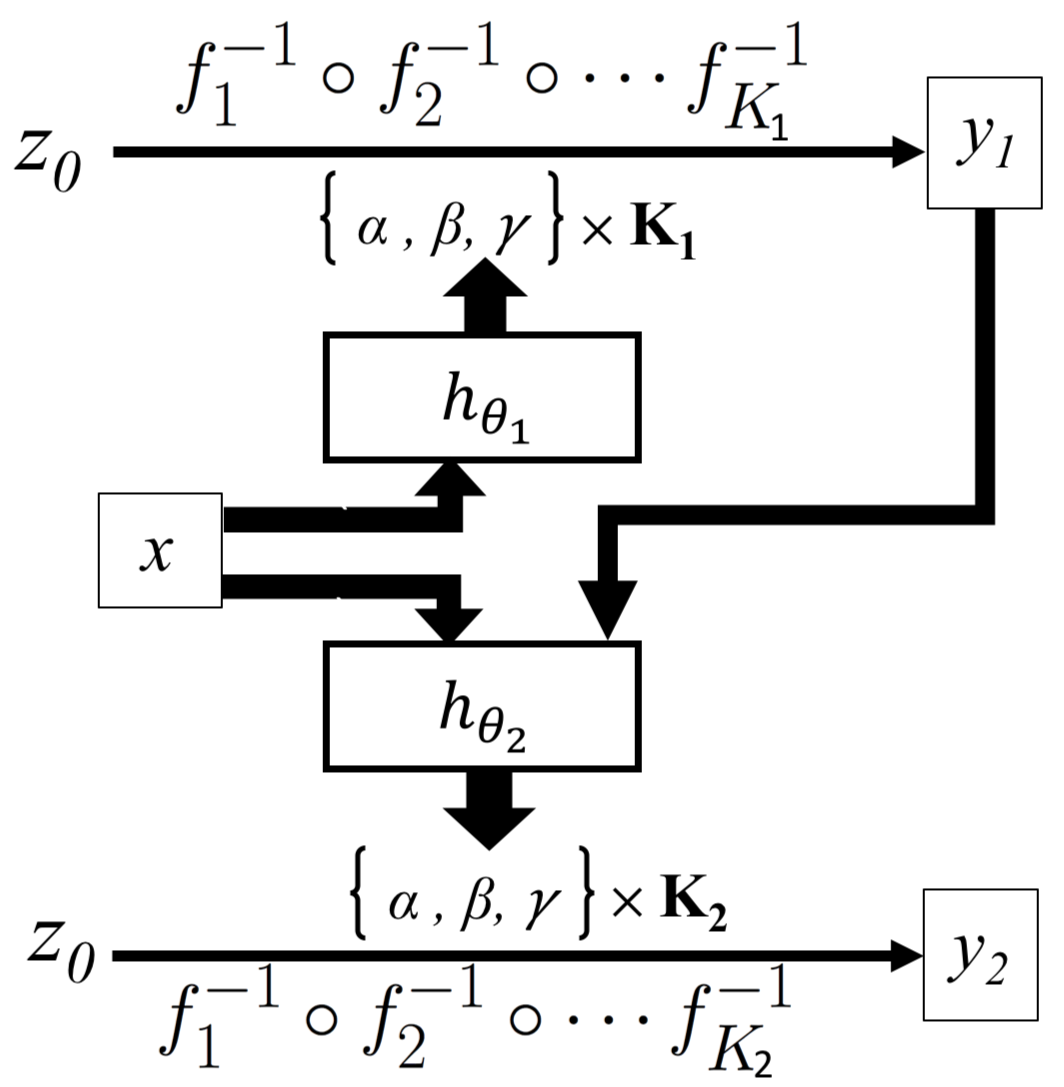}}
\caption{Schematic of conditional density estimation for a two dimensional predictive distribution using normalizing flows.  }\label{fig:nflow_cde_nade_model}
\end{center}

\end{figure}

\vspace{-5pt}
Up to this point, we have discussed only one dimensional prediction problems, but a number of applications of conditional density estimation, including spatial density estimation, demand higher dimensional predictive distributions.  In preliminary work, we investigated multidimensional radial and planar flows and variants thereof but were unable to successfully optimize these models to capture even relatively simple 2-dimensional densities.  To circumvent this challenge, we follow \cite{Larochelle2011} in using an auto-regressive structure to capture higher dimensional distributions.  In particular, by using the chain-rule of probability we can translate a 2-dimensional prediction problem into two one-dimensional problems as:
$$
p(y|x,\theta)=p(y_1|\omega_1=h_{\theta_1}(x))p(y_2|\omega_2=h_{\theta_2}(x,y_1))
$$
where $\omega_1$ and $\omega_2$ are the parameters of two 1D normalising flows.  This approach requires the use of two functions, $h_{\theta_1}: x \rightarrow \omega_1$, whose output is the parameters of $p(y_1|x)$, and $h_{\theta_2}: (x, y_1) \rightarrow \omega_2$, whose output is the parameters of the conditional, $p(y_2|y_1,x)$ (Supplementary Figure \ref{fig:nflow_cde_nade_model}).  As before, we choose to implement $h_{\theta_1}$ and $h_{\theta_2}$ as variational Bayesian neural networks.

\subsection{NYC Yellow Taxi Dataset} \label{sec:spatial}
\vspace{-5pt}
We applied our method to the NYC yellow taxi dataset which consists of more than 1 million trip records.  We performed variational inference as described in section \ref{sec:inference} in a model predicting a distribution over the taxi pickup locations given the fare of the ride, the percent tip, the time of day and the number of passengers.  We normalized the distribution of each feature to be zero-mean with unit variance, and when making predictions conditioned on just a subset of these features, we approximately marginalized over the other features by sampling several values of the missing dimensions from a unit normal and averaging the predictive distributions.

To accommodate the complexity of the conditional distributions and precisely capture subtle changes with respect to time of day, fare amount and percent tip, we implemented both $h_{\theta_1}$ and $h_{\theta_2}$ as neural networks, each with $2$ layers of $200$ hidden units and defined each density using a $20-$stage normalizing flow.  Additional details are included in supplementary section \ref{sec:nyc_details}.

This approach allows us to derive several insights about taxi pickups in Manhattan.  For example, we observe several trends by looking at the conditional densities for different tip and fare amounts (Figure \ref{fig:taxi_fare_tip}).  For $\$5$ fares, we see an increased density in the upper East side, a wealthy, primarily residential area.  This density is notably larger for trips with $20\%$ tips than rides with no tip.  Rides with no recorded tip with $\$5$ or $\$10$ fares are most highly concentrated in midtown around Times Square and along $5^{\mathrm{th}}$ avenue, which are particularly touristy areas of the city.  For $\$20$ fares, density is higher around Wall Street, with notably higher density for trips with no tip than with a $20\%$ tip.

\vspace{-5pt}
\section{Conclusion}
\vspace{-5pt}
We have demonstrated an efficient method for using normalising flows as a flexible likelihood model for conditional density estimation.  To confront fundamental trade-offs between modeling distributional complexity, functional complexity and heteroscedasticity we introduced a Bayesian framework for placing priors over conditional density estimators defined using normalising flows and performing inference with variational Bayesian neural networks.  Normalising flows and Bayesian approaches present exciting directions in conditional density estimation.  We believe that future work on constructing interpretable priors for conditional density estimators and improved Bayesian inference schemes will lead to further improvement of the flexibility and power of these approaches.

\vspace{-5pt}
\section{Acknowledgements}
\vspace{-5pt}
The authors would like to thank Laurence Aitchison, John Bradshaw, Shakir Mohamed and Amar Shah for insightful comments and discussion.  BLT additionally thanks the Euretta J Kellett fellowship for financial support.

\bibliography{Mendeley}
\bibliographystyle{icml2017}

\clearpage
\section{Supplement}
\begin{table*}
\caption{Mean log-likelihood in nats for normalizing flows, mixture density networks and neural networks with latent inputs on six small UCI benchmark regression dataset.  Higher is better.}\label{table:uci_results_cde}
\begin{center}
\fontsize{6.2}{8.5}\selectfont
\begin{tabular}{|p{.70cm}|p{0.4cm}|p{0.20cm}|p{1.05cm}|p{1.05cm}|p{1.05cm}|p{1.05cm}|p{1.05cm}|p{1.05cm}|p{1.05cm}|p{1.05cm}|p{1.05cm}|p{1.05cm}|}
\hline
\textbf{Dataset} & N & D & \textbf{MDN-2} & \textbf{MDN-5} & \textbf{MDN-20} & \textbf{LV-15} & \textbf{LV-5} & \textbf{NF-2} & \textbf{NF-5} &  \textbf{HMC} & \textbf{Dropout} & \textbf{MF}\\
\hline
boston & 506 & 13 & -2.65$\pm$0.03 & -2.73$\pm$0.04 & -2.74$\pm$0.03& -2.64$\pm$0.05 & -2.56$\pm$0.05 & -2.40$\pm$0.06 & -2.37$\pm$0.04 &\textbf{-2.27}$\pm$\textbf{0.03}& -2.46$\pm$0.25 &-2.62$\pm$0.06\\
concrete & 1030 & 8 & -3.23$\pm$0.03 & -3.28$\pm$0.03& -3.27$\pm$0.02 & -3.06$\pm$0.03 & -3.08$\pm$0.02&-3.03$\pm$0.05&-2.97$\pm$0.03&\textbf{-2.72}$\pm$\textbf{0.02}&-3.04$\pm$0.09&-3.00$\pm$0.03\\
energy & 768 & 8 & -1.60$\pm$0.04 & -1.63$\pm$0.06 & -1.48$\pm$0.04 & -0.74$\pm$0.03 & -0.79$\pm$0.02 & \textbf{-0.44}$\pm$\textbf{0.04} & -0.67$\pm$0.15 & -0.93$\pm$0.01 & -1.99$\pm$0.09 & -0.57$\pm$0.04\\
power & 9568 & 4 & -2.73$\pm$0.01 & -2.70$\pm$0.01 & -2.68$\pm$0.01 & -2.81$\pm$0.01 & -2.82$\pm$0.01 & -2.73$\pm$0.01 & \textbf{-2.68}$\pm$\textbf{0.01} & -2.70$\pm$0.00 & -2.89$\pm$0.01 & -2.79$\pm$0.01\\
wine & 1588 & 11 & -0.91$\pm$0.04 & \textbf{1.43}$\pm$\textbf{0.07}&1.21$\pm$0.06 & -0.98$\pm$0.02 & -0.96$\pm$0.01 & -0.87$\pm$0.02 & -0.76$\pm$0.10 & -0.91$\pm$0.02 & -0.93$\pm$0.06 & -0.97$\pm$0.01\\
yacht & 308 & 6 & -2.70$\pm$0.05 & -2.54$\pm$0.10 & -2.76$\pm$0.07 & -1.01$\pm$0.04 & -1.15$\pm$0.05 & -0.30$\pm$0.04 & \textbf{-0.21}$\pm$\textbf{0.09} & -1.62$\pm$0.02 & -1.55$\pm$0.12 & -1.00$\pm$0.10\\
\hline
\end{tabular}
\end{center}
\end{table*}

\subsection{Comparisons}\label{sec:comparisons}
In this supplementary section we report implementation details of the models evaluated in section \ref{sec:benchmarks}
In all models, $h_\theta$ is defined to be an MLP with a single hidden layer of $50$ units with tanh activations.  We optimize with Adam \cite{Kingma2015a} with hyper parameters $\beta_1=0.9$ and $\beta_2=0.99$ and learning rate $0.005$ and ran batch optimization for $5000$ iterations.  In all models, we use the local reparameterization trick \cite{Kingma2015} and calculate the $D_{\mathrm{KL}}(q||p)$ and its gradient analytically.  All models use $20$ MC samples of weights and biases during both training and testing.

For both the MDN and LV models, we placed zero-mean Gaussian priors on the weights and biases. In previous work, we found that the mean field approximations were sensitive to the prior standard deviation, so optimized this hyperparameter using Bayesian optimization.  Following \cite{Blundell2015} and our personal experience, we found better performance initializing the variance of the approximate posterior over weights to be very small ($\sigma^{\mathrm{init}}=10^{-5}$).  This seems to allow the models to fit to data before incurring the strong regularization penalty imposed by the entropy term in the variational free energy.

\subsubsection{Mixture density networks}
We follow \cite{Papamakarios2016} in using stochastic VI to implement a Bayesian mixture density network (MDN).  We use a diagonal Gaussian approximate posterior over weights and biases.  We test three models, with $2$. $5$ and $20$ mixing components which we referred to as MDN-2, MDN-5 and MDN-20, respectively.

Breaking from previous implementations of MDNs, we parameterized the model likelihood with an additional offset parameter, $s$, such that $\omega=\cup_{c=1}^C\{\mu_c, \sigma_c, \lambda_c\} \cup \{s \} $:

$$
p(y|\omega)= \sum_{c=1}^C\lambda_c \mathcal{N}(y|\mu_c+s, \sigma_c^2)
$$

where we enforce $\sum_{i=1}^C \lambda_c =1$ by defining $\lambda = \mathrm{Softmax}(\hat \lambda)$, and enforce $\sigma_c >0$ by defining $\sigma_c=\mathrm{softplus}(\hat \sigma_c)$, where $\hat \lambda$ and $\hat \sigma$ are unconstrained outputs of the neural network.  This parameterization is redundant in two ways, first, the global shift given by $s$ could equivalently be encoded by shifts in the means each of the mixing components and second, $\lambda$ is a $C$-dimensional simplex variable with only $C-1$ degrees of freedom.  We prefer this parameterization of the component means because, in the context of our variational neural network approximation, it does not penalize global shifts distributions according to the complexity of these distributions (as would be the case otherwise).  As such, a complex but homoscedastic noise distribution may be more easily represented.  In total, a mixture density network constructed in this way with $C$ mixing components will have $3C+1$ outputs.

\subsubsection{Bayesian Normalizing Flows}
We used a diagonal Gaussian variational approximation with fixed weight variances as this model performed well in the model assuming homoscedastic, Gaussian observations.  We tested two normalizing flow based conditional density estimators with different levels of complexity, one with two radial warpings (NF-2) and one with five radial warpings (NF-5).  We chose these models to have similar expressive power and the same number of parameters as the two mixture density network models, MDN-2 and MDN-5.  We performed hyperparameter optimization on three of the models' hyper parameters, $\sigma_{\hat \beta}$, $\lambda$, which set priors over the maximum changes of log probability density relative to the base distribution and the amount of heteroscedasticity, respectively as well the posterior variances of the weights and biases  connecting into the hidden layer, $\sigma_w$.  We a unit normal prior over the biases for the $\gamma$s and used a normal prior over biases for the $\alpha$s with mean 1 and variance 1, reflecting a prior that the predictive distributions would be relatively smooth (Figure \ref{fig:sampled_distributions}).  In addition to the radial flow parameters, we additionally predict a global translation of the noise distribution.  This can be thought of as a final stage of the normalising flow with slope 1 and an input dependent offset.  In total a conditional density estimator with Bayesian normalising flows with $K$ radial flows will have $3K+1$ outputs.

\subsubsection{Neural Networks with Latent Variables}
The third conditional density estimator which we tested is a neural network with latent variable inputs. We fit this model by mean field variational inference, following the implementation used by \cite{Depeweg2016}.  We tested two models, one in which we calculated likelihoods using $5$ samples of noise(LV-5) and one calculating likelihoods with $15$ samples (LV-15).  We make this choice attempting to pick models with roughly the same expressive power as the corresponding normalizing flow and MDN models, however, given the marked difference between the approaches, an objective comparison of expressive power is not possible.  As with the Bayesian MDN's, we initialized the standard deviations of the approximate posteriors to be $10^{-5}$ and selected the prior standard deviation using Bayesian optimization. 

Unlike \cite{Depeweg2016}, we use tanh activation units.  Previous work used ReLU hidden units which, when using input noise sampled from a uniform distribution, results in densities which are piece-wise linear.  As such, learning such, we found learning distributions with small ReLU networks in this way to be very difficult as compared to similar networks with smooth activation functions.

This approach is additionally more difficult to scale up in a stochastic variational Bayesian framework, where we are forced to use Monte Carlo samples over model parameters as well as the inherent stochasticity.  Similarly, evaluating the density under the posterior predictive distribution requires multiple samples for both weights of the network and the input noise.

\begin{equation}\label{eqn:bayesian_input_noise_approx}
\begin{split}
p(y|x,\mathcal{D},\alpha,\eta) &=   \int_\theta p(\theta|\mathcal{D},\alpha,\eta)  \int_z p(z|\eta) p(y|x,z , \theta)dz d\theta \\
&\approx \frac{1}{M} \sum_{i=1}^M \frac{1}{K} \sum_{j=1}^K p(y|x,z_j, \theta_i)
\end{split}
\end{equation}
where each $\theta_i \sim p(\theta|\mathcal{D},\alpha,\eta)$ and each $z_j \sim p(z|\eta)$

We report the mean and standard error of performance on held-out datasets in table \ref{table:uci_results_cde}.

\subsection{Radial flow gradient derivation}
Here we provide the derivation of the log gradient of the radial flow as presented in equation \ref{eqn:new_flow_gradient}:

\begin{equation}\label{eqn:rad_flow_gradient}
\begin{split}
\frac{d f(z)}{dz} &= 1+\frac{d}{dz}\frac{\alpha\beta r}{\alpha + |r|} \\
 &= 1+\frac{(\alpha + |r|)\frac{d}{dz}(\alpha\beta r)-\alpha\beta r \frac{d}{dz}(\alpha + |r|)}{(\alpha + |r|)^2} \\
&= 1+\frac{\alpha^2 \beta + \alpha \beta |r|- \alpha\beta|r|}{(\alpha + |r|)^2}\\
&= 1+\frac{\alpha^2\beta}{(\alpha + |r|)^2}
\end{split}
\end{equation}

where $r=z - \gamma$.  

\subsection{Spatial conditional density estimation experimental details on the NYC yellow taxi dataset}\label{sec:nyc_details}
As our observed variables, we consider pickup time, number of passengers, fare amount, and percent tip.  We presented time of day as an input to the model, parameterizing with two variables as $\big(\mathrm{sin}(\frac{2\pi\cdot\mathrm{hour}}{24}),\mathrm{cos}(\frac{2\pi\cdot\mathrm{hour}}{24})\big)$.  This parameterization of time is preferable in that it ensures that similar times are close in input space (e.g. 11:59pm is close to 12:00am, which is not the case for a one dimensional parameterization of time).

All input features and labels are normalized to have zero-mean a unit variance.  We performed variational inference using the local reparameterization trick.  We used a learning rate of $2\cdot 10^{-5}$, and ran for $1500$ epochs with batch size of $2048$ on a NVIDIA Tesla K80 GPU. For hyper-parameters, we set $\mu_{\hat \alpha}=1$ and $\sigma_{\hat \alpha}=0.1$, $\lambda=1$,$\mu_{\hat \beta}=0$ and $\sigma_{\hat \beta}=1$, $\mu_{z}=0$ and $\sigma_{z}=1$, and the prior over weights to be unit Gaussian.  We fixed the posterior uncertainties in weights and biases to $10^{-5}$.

We use data made publicly available \href{http://www.nyc.gov/html/tlc}{www.nyc.gov/html/tlc}.  We used the data from January of 2016, as it provided an interesting proof of principle and scaling up to the entire dataset would have posed challenges outside the scope of this work.

\begin{figure*}[!ht]
\begin{center}
\centerline{\includegraphics[width=0.6\textwidth]{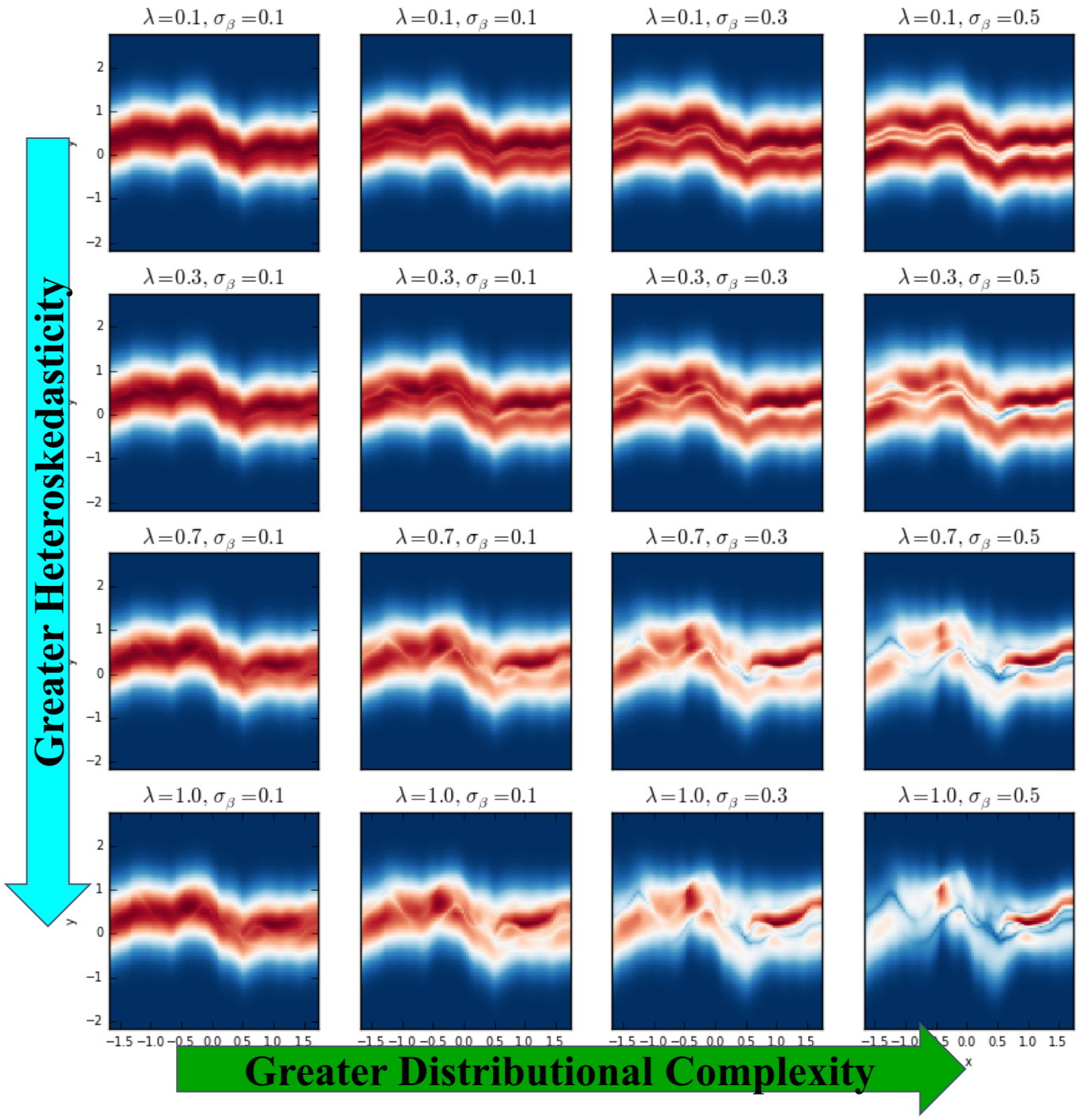}}
\caption{An additional slice of the manifold in Figure \ref{fig:cd_manifold_mid_ls} with a shorter length scale.}\label{fig:cd_manifold_fast_ls}
\end{center}

\end{figure*}

\subsection{Toy Results}
\begin{figure*}[!ht]
\begin{center}
\centerline{\includegraphics[width=0.9\textwidth]{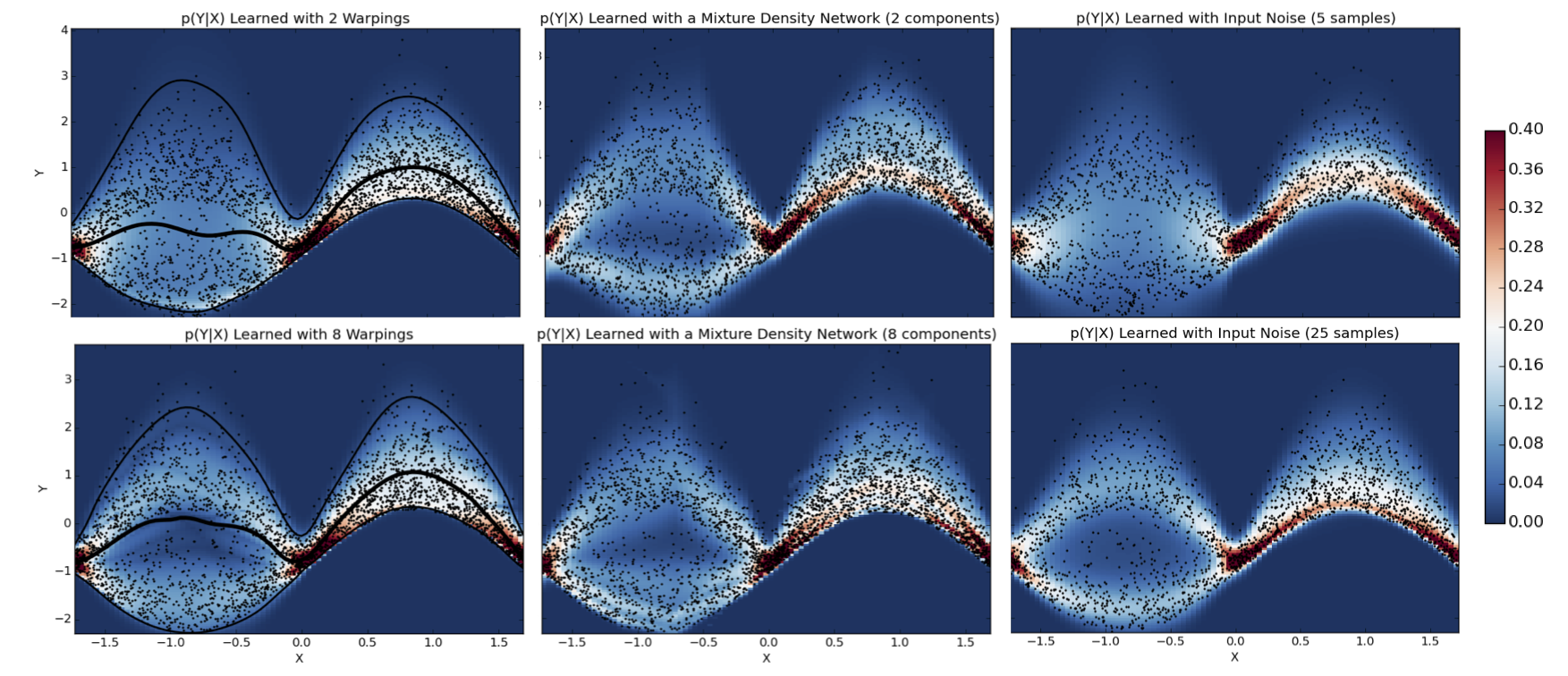}}
\caption{Heatmaps representing toy densities learned with three conditional density estimation methods. LEFT) Normalizing flows, MIDDLE) Mixture density networks and RIGHT) input noise.  The top row demonstrates the performance of simple models and the bottom show performance for higher capacity models.  Heat represents the conditional probability, $p(y|x)$.  Normalizing flows additionally allow us easily find confidence intervals of this conditional distribution; we plot the $95\%$ confidence interval in black.  The training set consisted of $5000$ points.  Best viewed in color.}
\label{fig:cde_toy_heatmaps}
\end{center}
\end{figure*}

In this section we present results fitting three methods with complex likelihood models to toy data.  The three models tested are normalizing flows, mixture density networks and neural networks with latent variables.  For all models we used a multilayered perceptron with a single hidden layer with $50$ units to set the input dependence.  For the latent input model we included $5$ noisy inputs, from which we took 20 Monte Carlo samples to approximate a likelihood (as done in the inner sum of equation \ref{eqn:bayesian_input_noise_approx}).  The learned densities are shown in figure \ref{fig:cde_toy_heatmaps} with numerical performance given in table \ref{table:toy_results}.  In all of these models, we used Xavier weight initialization, do not impose and priors or regularization and fit by batch gradient descent.

\begin{table}
\caption{Held-out mean log-likelihood in nats for normalizing flows, mixture density networks and neural networks with latent inputs on a toy regression task (N=5000).  The simple and complex models refer to the two levels of complexity described in section \ref{sec:comparisons}.  Higher is better.)}\label{table:toy_results}
\begin{center}
\fontsize{10}{14}\selectfont
\begin{tabular}{ |p{1.5cm}||p{1.7cm}|p{1.7cm}|p{1.7cm}|  }
 \hline
 \multicolumn{4}{|c|}{} \\
 \hline
Method & Normalizing Flows & Gaussian Mixture & Latent Inputs \\
 \hline
 Simple   & 2.12    & 2.10 &   2.25\\
  \hline
Complex &   2.08  & 2.07   & 2.16 \\

\hline
\end{tabular}
\end{center}
\end{table}

\subsection{A note on non-parametric conditional density estimators}
Perhaps the best alternative to the approach we have taken for conditional spatial density estimation is nonparametric conditional density estimation, which can be used to probe the observed distribution of events with certain characteristics.  This approach has several disadvantages compared to parametric approaches.  In particular, nonparametric methods require passing through the entire dataset for each prediction and demand one to set potentially arbitrary bounds to filtering points to be included in calculation of conditional densities \cite{Bashtannyk2001}.  In contrast, though they may require many passes through the dataset during training time, parametric methods make constant time predictions which interpolate between the densities where data has been observed, avoiding the need to set bounds for filtering data.

\subsection{Chicago Crime Modeling}\label{sec:chicago_crime}
\begin{figure*}[!ht]
\centering
\includegraphics[width=0.9\textwidth]{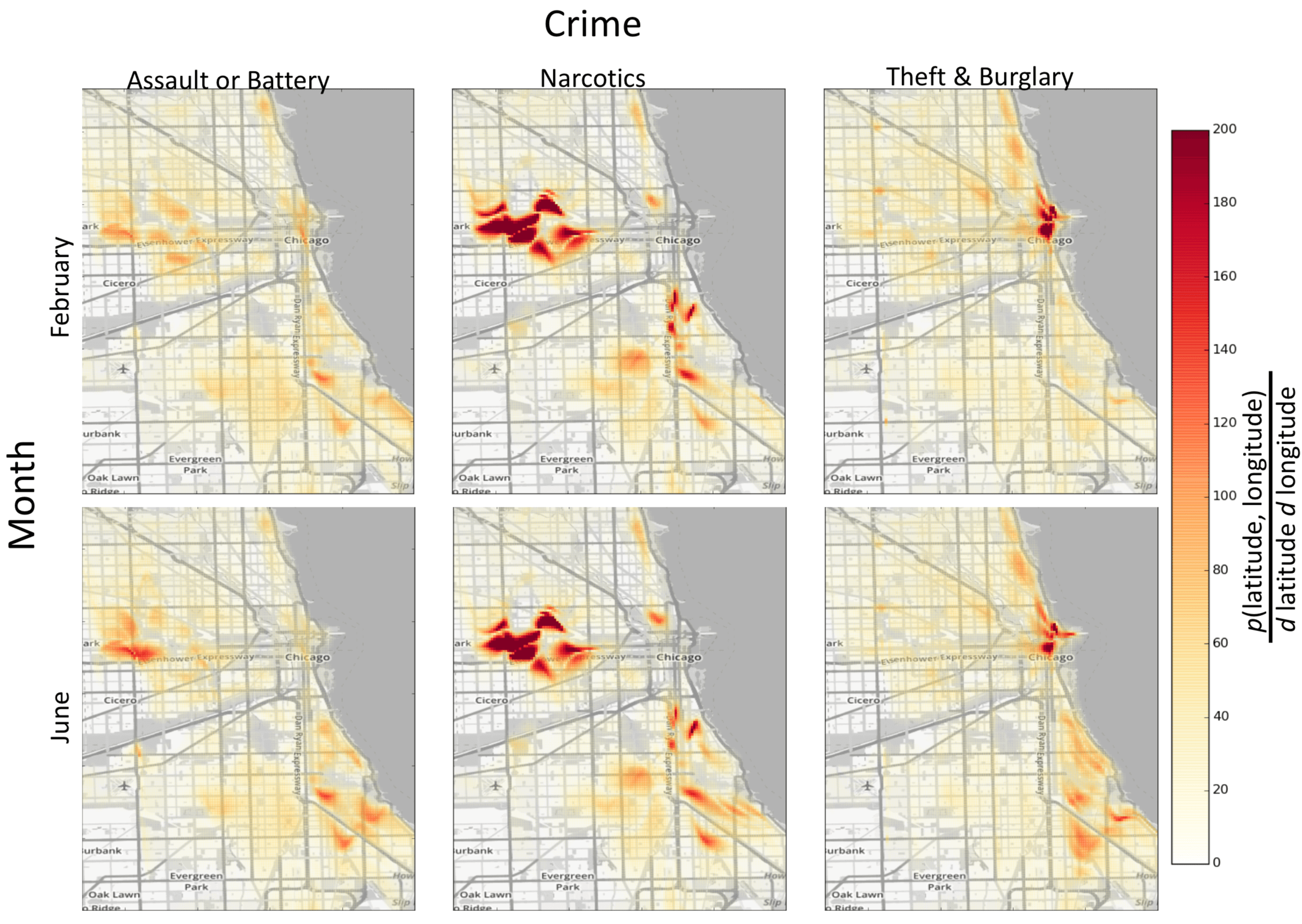}
\vspace{-9pt}
\caption{Heatmaps representing learned conditional densities of crime reports throughout Chicago in the winter and summer, broken down by crime type.  Heat density is in units of $\frac{p(\mathrm{lat},\mathrm{long})}{d \mathrm{lat}\ d\mathrm{long}}$ and is capped at $1000$.  Best viewed in color.}\label{fig:chicago_crime}
\vspace{-9pt}
\end{figure*}

We additionally applied our method to model conditional distribution of the location of reported crimes in Chicago\footnote{Data available from https://data.cityofchicago.org/Public-Safety/Crimes-2001-to-present/ijzp-q8t2}.   Supplementary figure \ref{fig:chicago_crime} demonstrates the differences in the distribution of several different types of crime in the winter and summer.  One can see that the density shifts dramatically for different classes of violations, and exhibits shifts across the locations of these densities.   As with the yellow taxi dataset, we implemented both $h_{\theta_1}$ and $h_{\theta_2}$ as neural networks with $2$ layers of $200$ hidden units and defined each density using a $20-$stage normalizing flow.

\begin{figure*}
\begin{center}
\centerline{\includegraphics[width=0.9\textwidth]{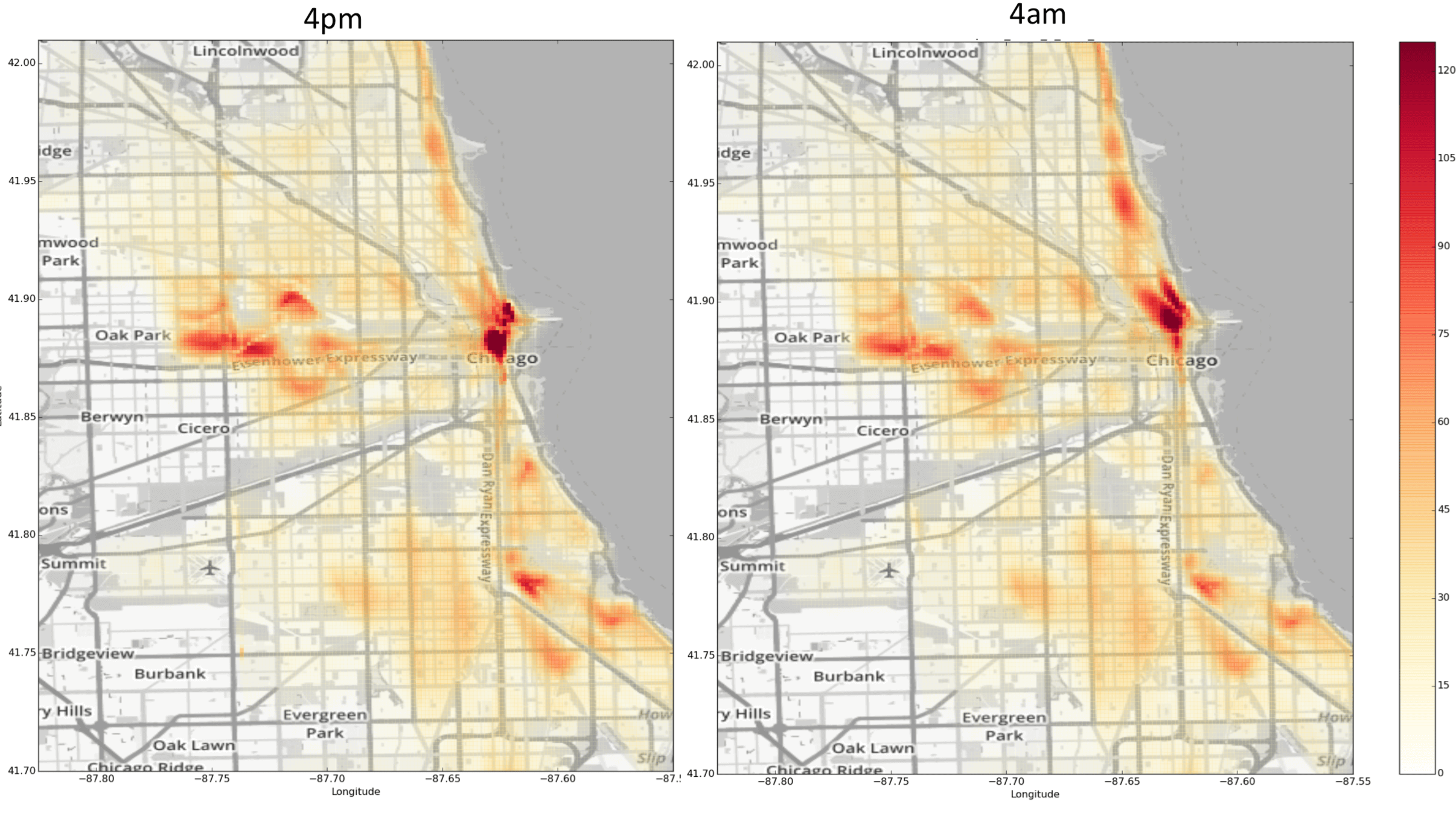}}
\caption{Crime density in chicago at different times of day.}
\label{fig:chicago_time}
\end{center}
\end{figure*}

\end{document}